# Cohesion-6K: An Arabic Dataset for Analyzing Social Cohesion and Conflict in Online Discourse

**Aisha Ali Al-Athba, Wajdi Zaghouani**
Hamad Bin Khalifa University, Northwestern University in Qatar
Doha, Qatar
aisha.alathba@hbku.edu.qa, wajdi.zaghouani@northwestern.edu

**Abstract**

The study of online discourse has become central to understanding societal polarization. While much research has focused on detecting overt toxicity, the subtle dynamics of social cohesion, meaning the interaction between divisive and unifying narratives, remain computationally underexplored (Bail, 2021; González-Bailón and Lelkes, 2023). This paper presents *Cohesion-6K*, a manually and ChatGPT-assisted annotated dataset of six thousand Arabic public Facebook posts related to the Israeli Occupation of Palestine. Each post is assigned to one of five discourse categories that represent a continuum from conflict to cohesion: *Conflict*, *Resolution*, *Community Engagement*, *Supportive Interactions*, and *Shared Values*. The annotation process combines expert human judgment with model-assisted pre-labeling verified by trained annotators, achieving substantial inter-annotator agreement (Cohen's $\kappa = 0.85$). Quantitative analysis reveals a consistent engagement gap, where conflict-oriented posts receive between two and four times more user interaction than resolution-oriented ones ($p < 0.01$). This pattern illustrates how divisive discourse tends to attract disproportionate visibility in Arabic social media spaces. *Cohesion-6K* provides a transparent and reproducible resource for the study of online cohesion and polarization. The dataset, annotation guidelines, and preprocessing code will be released for research use under an open license, supporting future work in computational social science, digital communication, and Arabic natural language processing.

## 1. Introduction

### 1.1. Research Context

The Israeli Occupation of Palestine stands as one of the most enduring and complex geopolitical issues of the modern era, serving as a constant focal point for global media and public discourse. The advent of social media has radically transformed the landscape of political engagement surrounding this conflict. Platforms like Facebook, X (formerly Twitter), and TikTok enable diverse narratives and interactions that transcend geographical barriers, fundamentally reshaping the dynamics of social cohesion and division (González-Bailón and Lelkes, 2023; Suwarno and Sahayu, 2020). The digital realm has evolved into both a battlefield of ideologies and a space for unprecedented community building and advocacy, making it a critical area for scholarly investigation.

Scholarship defines social cohesion as a network of social interactions that promote collective unity (Terren and Borge-Bravo, 2021), built on shared values and trust. In the digital age, these interactions are increasingly mediated by platforms that shape how communities form, interact, and sustain collective actions. However, the literature reveals a complex dichotomy: while these platforms can foster understanding and solidarity, they also serve as potent arenas for misinformation, ideological polarization, and emotional manipulation.

Our research provides significant insights into this dynamic, demonstrating how conflict-driven content consistently receives more engagement than peace-oriented initiatives, a trend that exacerbates ideological polarization and undermines efforts toward reconciliation.

### 1.2. Research Problem

Despite extensive studies on social media's role in public discourse, a significant gap persists in understanding its specific impact on social cohesion within the context of the Israeli Occupation of Palestine. Much of the existing research focuses on short-term case studies or platform-specific analyses without adequately addressing the long-term evolution of digital narratives. For instance, narratives shifted dramatically following key events like the Sheikh Jarrah evictions, but the underlying mechanisms of community engagement and cohesion remain poorly understood.

This evolving digital discourse creates a complex landscape of both unity and division, yet there is a lack of structured, annotated data to computationally analyze these dynamics. Without such resources, it is difficult to move beyond anecdotal observations to empirically measure the prevalence and impact of different narrative types. This research aims to fill this gap by creating and analyzing a dataset to dissect the intricate relationship between social media narratives and their impact on community dynamics in a highly polarized environment. Understanding these dynamics is crucial for developing effective communication strategies and platform interventions that can mitigate conflict and promote social cohesion through digital channels.

## 1.3. Research Questions and Objectives

This study is guided by a central research question: how do various cultural, psychological, and algorithmic factors shape the engagement with digital posts related to the Israeli Occupation of Palestine, and how do these interactions contribute to social cohesion and division in online spaces? To address this question, this paper focuses on the following objectives:

1. To develop a comprehensive annotation framework for classifying social media content based on its potential to foster or undermine social cohesion.
2. To create and validate a large-scale, annotated dataset of Facebook posts representing diverse perspectives on the conflict.
3. To quantitatively analyze engagement patterns across different discourse categories to identify which narratives are amplified.
4. To compare the efficacy of manual versus hybrid AI-assisted annotation methodologies for this complex task.

## 1.4. Structure of the Paper

This paper is structured to provide a comprehensive exploration of our dataset and findings. Section 2 reviews the relevant literature on social cohesion, digital discourse, and conflict. Section 3 details the methodology, including our data collection process, annotation guidelines, and reliability measures. Section 4 presents the core research findings, including the distribution of content categories, an analysis of engagement patterns, and a comparison of annotation methods. Section 5 discusses the broader implications of these findings for social cohesion and platform governance. Finally, Section 6 concludes the paper, summarizing its contributions, acknowledging its limitations, and proposing directions for future research.

# 2. Related Work

This research is informed by several interconnected domains: theoretical frameworks of social cohesion in digital spaces, the role of social media in political conflicts, platform-specific discourse analyses, and the psychological dimensions of online engagement.

## 2.1. Theoretical Frameworks

### 2.1.1. Social Cohesion in Digital Spaces

Traditional studies define social cohesion as a society's ability to maintain supportive networks built on shared values and trust. The digital age has profoundly transformed these interactions. Online spaces are characterized by unique dynamics such as anonymity, rapid information flow, and algorithm-driven content visibility, which create new challenges and opportunities for social cohesion. During crises, these platforms can empower marginalized communities and facilitate collective action. However, they also risk fostering polarization and fragmentation, particularly in highly contentious contexts like the Israeli-Palestinian conflict.

### 2.1.2. Schwartz's Cultural Value Theory

Schwartz's Cultural Value Theory provides a powerful lens for understanding how shared values shape online political communication (Russo et al., 2022). The framework identifies key values such as security, power, and benevolence as primary drivers of interaction (Russo et al., 2022).

Security-oriented users often prioritize national defense and identity, leading them to amplify nationalist posts that reinforce in-group solidarity while isolating opposing views. Power-driven users may use social media to dominate discussions, often employing tactics that suppress dissent and contribute to ideological polarization. Benevolence-focused users strive to foster unity and bridge divides; however, their efforts often struggle against algorithmic biases that favor divisive and emotionally charged content.

### 2.1.3. Agenda-Setting and Framing Theories

Agenda-setting theory posits that media does not merely reflect reality but shapes it by selecting which issues to highlight. On social media, this process is driven by trending algorithms, viral hashtags, and content curation, which collectively mold user perceptions of the conflict (Kareem and Najm, 2024). Framing theory complements this by focusing on how information is presented to elicit specific emotional responses. Conflicts are often framed in ways that provoke strong emotional reactions like moral outrage, which fosters deeper ideological divisions (Tao et al., 2024).

### 2.1.4. Misinformation and Digital Polarization

Echo chambers are digital environments where users are primarily exposed to reinforcing beliefs and therefore represent fertile ground for misinformation. Unverified and sensational content spreads rapidly, appealing to users' cognitive biases (Muhammed and Mathew, 2022). The propagation of false narratives erodes trust in established institutions and media, skewing public perception and deepening ideological divides (Terren and Borge-Bravo, 2021). Over time, this dynamic fosters extreme partisanship, where individuals become so entrenched in their views that they react adversely to balanced or objective information, ultimately harming societal cohesion.

## 2.2. Social Media and Political Discourse in Conflicts

Social media serves a dual role in conflicts: it is both a powerful tool for activism and a conduit for propaganda. It allows marginalized voices to gain global visibility through viral hashtags, but it is also used by state actors and political groups to craft manipulative narratives using emotional appeals and selective information (Foos et al., 2021). This discourse is heavily shaped by political leaders and media outlets who use their platforms to direct public discussion in line with their strategic interests. During crisis events, user activity spikes as people seek information and express support. This heightened engagement often fuels polarization, as rapid, emotional reactions are prioritized over analytical thinking, deepening societal divisions (Arrosvid and Halwati, 2021). Furthermore, the battlefield has expanded to include AI and automated bots, which are used to manipulate discussions by amplifying specific viewpoints and creating a skewed perception of consensus, posing a direct threat to authentic dialogue and social cohesion (Khaund et al., 2021).

## 2.3. Platform-Specific Analyses of Political Discourse

Different platforms shape political discourse in unique ways. X (formerly Twitter) facilitates rapid, real-time engagement through short posts and viral hashtags like #FreePalestine (Raza et al., 2023). Its algorithm, however, tends to prioritize engagement over accuracy, often promoting controversial content that may not facilitate nuanced discussion (Renhoran and Setiawan, 2024). Facebook and Instagram enable more detailed, visually rich storytelling through groups, pages, and personal stories (Raza et al., 2023), which can foster strong community bonds but also creates echo chambers where users see content that reinforces their pre-existing beliefs (Zahoor and Sadiq, 2021; Weninggalih and Pramiyanti, 2025). TikTok leverages short-form video and a powerful recommendation algorithm to engage younger demographics. Its format prioritizes emotional impact and rapid dissemination, often at the expense of factual accuracy, making it fertile ground for misinformation (Renhoran and Setiawan, 2024; Stern and Shalom, 2021).

## 2.4. Arabic NLP Resources for Social Media Discourse

The construction of annotated Arabic social media corpora has grown steadily in recent years, yet resources targeting political discourse and social cohesion remain scarce. Prior work has produced datasets addressing related phenomena, including hate speech (Zaghouani et al., 2024), stance detection (Charfi et al., 2024), and multi-dimensional sentiment and emotion analysis in politically charged contexts (Laabar and Zaghouani, 2024). Annotating such data presents well-documented challenges: the emotional intensity of the content imposes a measurable toll on annotators, and maintaining consistency across culturally loaded expressions requires extensive calibration (Al Emadi and Zaghouani, 2024). These concerns directly motivated the quality control procedures adopted in the present study.

Work on digital framing of the Israeli-Palestinian conflict in Arabic social media has begun to emerge. Shestakov and Zaghouani (2024) constructed a dataset focused on the digital framing of the Sheikh Jarrah evictions, demonstrating that event-specific corpora can reveal how platform dynamics shape narrative dominance. Our work builds on this trajectory by expanding the temporal and thematic scope, introducing a five-category cohesion-conflict taxonomy, and validating a hybrid human-AI annotation pipeline that Biswas et al. (2023) have shown to be promising for large-scale Arabic annotation tasks.

## 2.5. Gaps in the Literature

The existing literature reveals several critical gaps. There is an insufficient understanding of how specific algorithmic mechanisms influence social cohesion in politically charged environments. There is also a notable lack of longitudinal studies that track how digital discourse evolves in response to real-world events over extended periods. Finally, the influence of AI-driven content manipulation on social media narratives remains an under-researched area. This study begins to address these gaps by providing a structured, annotated dataset that serves as a foundational resource for such investigations.

# 3. Methodology

To systematically capture the multifaceted nature of online discourse, this study adopts a mixed-methods approach that integrates quantitative content analysis with qualitative thematic coding. This comprehensive strategy allows us to move beyond surface-level engagement metrics to conduct a nuanced analysis of thematic and ideological patterns (Braun and Clarke, 2021).

### 3.1. Research Approach

The core of our methodology is an integrated mixed-methods content analysis, which is particularly well-suited for examining complex debates on platforms like Facebook (Neuendorf, 2017). This approach allows for the exploration of both tangible indicators of engagement (likes, shares) and more subjective aspects like thematic undercurrents and ideological expressions. According to Creswell and Creswell (2018), combining quantitative and qualitative methods enhances the ability to interpret digital discourse, providing a richer understanding of both the extent and the depth of online interactions. Our approach is also interdisciplinary, incorporating insights from computer science through the use of machine learning-assisted annotation alongside manual data labeling to ensure both reliability and interpretive depth.

### 3.2. Data Collection and Dataset Composition

The dataset was collected from publicly accessible Arabic Facebook pages and verified accounts using the CrowdTangle platform, following Meta's data access and privacy policies. All posts are written primarily in Arabic, reflecting discourse surrounding the Palestinian-Israeli conflict and occupation from 1948 to the present, with a particular focus on the escalation beginning on 7 October 2023.

**Search Strategy and Keywords.** We employed a keyword-based retrieval strategy designed to capture Arabic digital discourse related to the conflict. The following search terms were used in the CrowdTangle interface:

ل""J:-\J|, "□ؚ□J٘J| □A,,aJ|", "ة:,□ عLk.'", "□□',,.ﷺJ| ڈسl.J|", تL.:13:...ﺀ□ا, "ةد3..J| J>-", "نL...□)J| ق3l.-", "تا'L1::;\J| □□tؚ:|,,...)J|", "□□\□k....tAJ| □,.وLl.ﺀ□ا", "تLفv;; ,,...L□", د□ سL½;;", "r..3□.J□J| وانv-", "□□tؚ:|,,...)J| تL4,,:::.at|", ةدL:إ, "Jπؚ:|,,...)J| ذوان..J|", "Jπؚ:|,,...)J| رL.,a>-J|", "r.٥٦J..J| □:lJ", "_;.....:J| لLl.:;;\J|", "ي,,.'''' .Jj□1;", "<..:ار\J| تا'L1::;ا □□\□k. tAJ|".

The Arabic search terms above correspond to: Occupation, West Bank, Gaza Strip, East Jerusalem, Settlements, Right of Return, Human Rights, Israeli Violations, Palestinian Resistance, Yasser Arafat, Mahmoud Abbas, Marwan Barghouti, Israeli Attacks, Genocide, Israeli Aggression, Israeli Blockade, Ethnic Killing, Arbitrary Detention, Forced Displacement, and Violations of Palestinian Territories.

**Temporal and Geographic Scope.** Posts were collected between 13 July 2022 (21:20 AST) and 28 May 2024 (21:20 AST). The country filter was left open to capture regional variation across Arabic-speaking communities and diasporic audiences.

**Sampling and Source Types.** The full data collection yielded approximately 200,000 Arabic Facebook posts retrieved through the CrowdTangle platform. From this corpus, a subset of 6,000 posts was selected for detailed manual and AI-assisted annotation. Sampling followed a stratified procedure designed to maintain diversity across source types (news outlets, NGOs, public figures, and grassroots movements) and engagement levels. Within each category, posts were randomly sampled to ensure proportional representation while controlling for duplication and spam. The 6,000-post limit was chosen to balance thematic coverage with the time and resources available for high-quality annotation and verification. To support reproducibility and future research, the larger unannotated corpus of approximately 200,000 posts will also be released for academic use under the same open license. Table 1 summarizes the dataset composition by source type and total engagement.

| Source Type | No. of Posts | Engagements |
|---|---|---|
| Political Organizations | 1,440 | 6.3 million |
| News Outlets | 2,070 | 9.1 million |
| Human-Rights Groups | 950 | 4.5 million |
| Public Figures | 860 | 3.7 million |
| Grassroots Movements | 680 | 2.8 million |
| **Total** | **6,000** | **26.4 million** |

Table 1: Composition of the *Cohesion-6K* Arabic Facebook dataset by source type and engagement.

## Illustrative Examples

To convey the linguistic and thematic diversity of the corpus, we present representative Arabic posts from the dataset followed by their English translations. All posts are drawn from the Arabic Facebook dataset collected through CrowdTangle.

**Example 1**

تL½:J|و .J□.,aJ| ،□½...,aJ| مL□\J| ه.k r./ ة:,□ □kأ □o:أ u,. Jπ.,a;
.i'C-J

*We pray for the people of Gaza in these difficult days; may you remain patient and steadfast.*

**Example 2**

.ᄉᄀ뽀ᄕ..JL: ᄕ̈ⓢ꿍 م""J...J' vلـلـ! J> Jk.J' ءاٴزL,.وـد LᅌA.:;; ;-

*Enough violence and destruction; the path to peace begins with justice.*

**Example 3**

.ل""J:-\J' مL,.أ ب4,.,a; uJ وL.:'3l.- ""\ئ uJ

*We will not forget our rights and we will not remain silent before the occupation.*

**Example 4**

.J٢l.J' ٣ 3,. r./ L.ၜ3' 3k r.ᄐJ..J' u,.L,,a:J'

*Arab solidarity is our strength in the face of oppression.*

These examples illustrate key themes that recur across the dataset, including expressions of solidarity, perseverance, justice, and resistance. Each instance reflects the communicative strategies used by Arabic speakers to express collective identity and emotional resilience during periods of political and humanitarian crisis.

**Ethical and Licensing Compliance.** Only publicly available posts were collected, and no personally identifiable information was retained. Metadata were anonymized and stored securely. The dataset will be released for research under a Creative Commons Attribution NonCommercial ShareAlike 4.0 license.

## 3.3. Annotation Scheme and Guidelines

The annotation framework for *Cohesion-6K* was developed to identify how Arabic online discourse expresses conflict and cohesion within the context of the Palestinian-Israeli situation. The scheme evolved through several iterative rounds of pilot annotation, discussion, and revision among three trained annotators who are native Arabic speakers with backgrounds in linguistics and media studies.

**Design of the Categories.** The taxonomy consists of five principal categories: *Conflict*, *Resolution*, *Community Engagement*, *Supportive Interactions*, and *Shared Values*. Each represents a functional dimension of social discourse rather than a sentiment label. The five categories were established through an iterative process of pilot annotation and discussion. The early version of the guidelines contained overlapping cues between *Conflict* and *Resolution*; these were refined after the first pilot to include pragmatic and lexical indicators unique to each category. For instance, *Conflict* posts typically employ accusatory verbs, negation, or adversarial framing, while *Resolution* emphasizes reconciliation, empathy, or a future-oriented tone. The final taxonomy reflects the consensus reached after multiple calibration sessions described in Section 3.4.

**Refinement and Calibration.** Annotation proceeded in three stages. During the first stage, annotators labeled one hundred Arabic posts independently and met to compare interpretations. Disagreements were discussed line by line to identify ambiguous expressions such as irony, coded language, or religious invocations. The second stage involved an updated guideline document with examples, followed by a new sample of five hundred posts. The inter-annotator agreement improved notably after these discussions. In the final stage, a consolidated manual was produced specifying linguistic cues, pragmatic indicators, and contextual exceptions. Annotators continued to meet weekly to resolve edge cases and record new observations for inclusion in future releases of the dataset.

**Illustrative Examples.** Below are representative examples from each category, accompanied by English translations. They illustrate the pragmatic and emotional range captured by the guidelines.

**Conflict**

.\k. tAJ' ,,al.J' ىL,.. u,. C ᅌL... uJ

*We will never forgive those who betrayed the Palestinian cause.*

**Resolution**

.ةL;L..ع' ءL1ሌ)J ذ-3J' J> Jk.J' 3k ىلL..J' م""J...J'

*A just peace is the only path to end the suffering.*

**Community Engagement**

.ة:, r./ o:L:ጠക' ,,...\J' r;;ᄊ뽀 ᄉᄀ뽀ᄊ/ ' r./ ١3-رLر

*Join the campaign to support families in need in Gaza.*

**Supportive Interactions**

.LᅌLᄖلـ L.:Lkى r./ 1:;أ ءr1;""JLkو vᄯJ-"J' L,. L._-'

*Our hearts are with the wounded and their families; you are always in our prayers.*

**Shared Values**

.ىL...! JSJ J>- ,.\J(-Jو J>-J'

*Freedom and dignity are rights for every human being.*

**Guideline Principles.** Annotators were instructed to focus on communicative intent and discourse function rather than lexical polarity. For example, a post criticizing violence can be coded as *Resolution* if it advocates dialogue rather than retaliation. Conversely, a post invoking faith or endurance may belong to *Supportive Interactions* when it provides comfort without political confrontation. To ensure uniformity, each guideline entry included examples, counter-examples, and notes on frequent sources of confusion such as sarcasm, rhetorical questions, and metaphorical expressions.

The final manual integrated the results of calibration sessions and contained explicit instructions for handling ambiguous tone, mixed sentiment, and code-switched text. Annotators achieved a stable understanding of category boundaries before the full annotation phase began. This iterative and participatory design ensured both cultural sensitivity and analytical reliability in the resulting corpus.

## 3.4. Annotation Process and Quality Control

To enhance the reliability and transparency of the annotation process, the 6,000 Arabic Facebook posts were subjected to a dual-stage classification framework involving both human expertise and AI assistance. The workflow was designed to balance interpretive nuance with large-scale efficiency, combining qualitative review cycles with quantitative reliability assessment.

**Dual-Stage Annotation.** The first 3,000 posts were labeled manually by a team of three trained native Arabic annotators. Each annotator received the finalized guidelines described in Section 3.3 and participated in multiple calibration sessions to ensure a shared understanding of category boundaries. The remaining 3,000 posts were pre-labeled using CHATGPT (GPT-4) and subsequently verified by the same human annotators. The prompt used for ChatGPT labeling instructed the model to assign each post to one of the five discourse categories based on the communicative intent of the text, following the category definitions and examples provided in the annotation guidelines. This hybrid approach combined the efficiency of AI-assisted labeling with the depth of human contextual reasoning, allowing a complete review of all labels before inclusion in the corpus.

**Quality Control Procedures.** Annotation quality was maintained through continuous monitoring and feedback. Twenty-five percent of the dataset (1,500 posts) was triple-coded by all annotators to evaluate consistency. Random samples from the remaining data were reviewed weekly by the lead researcher. Annotators recorded ambiguous or context-dependent items in a shared log for discussion during review meetings. Cases involving irony, sarcasm, religious expressions, or emotionally charged content were flagged for collective interpretation to refine the decision rules. All updates to the guidelines were versioned, and the evolution of definitions was documented to ensure reproducibility.

**Reliability Assessment.** Inter-Annotator Agreement (IAA) was computed using Cohen's $\kappa$, which corrects for agreement expected by chance. The resulting score of $\kappa = 0.85$ indicates a high level of consistency among human annotators. In the field of content analysis, values above 0.80 are generally considered excellent (Landis and Koch, 1977). This level of agreement confirms the reliability of the final dataset and the effectiveness of the iterative training and feedback procedures.

**Accuracy Evaluation.** To evaluate the accuracy of each annotation method, a gold-standard reference set was constructed from 300 posts drawn from the triple-coded portion of the dataset. These 300 posts had full three-way annotator agreement and were therefore treated as reliable ground truth. Manual annotation accuracy (92% on average) reflects the proportion of labels assigned by individual annotators that matched this gold standard across the non-triple-coded portion of the manually annotated set. AI-assisted accuracy before verification (84–87%) reflects ChatGPT's agreement with the gold standard on a 300-post held-out evaluation sample drawn from the AI-labeled half. AI-assisted accuracy after human verification (88–91%) reflects the final label quality of the AI-labeled set after annotators reviewed and corrected ChatGPT's outputs. These figures are reported per-category in Table 2.

### 3.4.1. Annotation Challenges and Disagreement Analysis

Despite the strong agreement rate, some linguistic and cultural nuances created recurrent annotation difficulties. These cases provided insight into the complexity of Arabic discourse and informed subsequent improvements to the guidelines.

**Conflict vs. Resolution.** Posts blending critique and calls for peace often generated disagreement.
[illegible]

*Peace will be achieved only when the occupation ends and rights are restored to their owners.*
Some annotators coded this as *Conflict* due to accusatory framing, while others labeled it as *Resolution* because of the explicit call for peace. The team decided to prioritize constructive intent as the determining factor, updating the guidelines accordingly.

**Religious and Moral Expressions.** Faith-based language often intertwined emotional support with political stance.

...\il نذ1- مدL' ,,..:2.:Jا ،□:<3Jا r..;و ..\il L.:□...-

*God is sufficient for us, and He is the best protector; victory will come, God willing.*
Annotators disagreed between *Supportive Interactions* and *Conflict*. After discussion, consensus favored *Supportive Interactions* when the message centered on hope and perseverance rather than hostility.

**Irony and Sarcasm.** Arabic irony frequently uses positive wording to deliver criticism.

!م3□ J' □:L.:l.Jا □t□□ يᄉᆸᄀᄇᄂا م""J...Jا ا.k □□뽀쁘 3k 꾮ơ

*How beautiful is this peace that brings bombs every day!*
Some annotators labeled it as *Resolution* literally, while others interpreted it as sarcastic and thus *Conflict*. The guideline was updated to treat sarcasm expressing disapproval or ridicule as *Conflict*.

**Shared Values vs. Supportive Interactions.** Moral declarations were occasionally confused with empathetic appeals.

.Lő..,. r.ᄀᄅL...□.و .'1و □واؤ u,.L,,a:Jا

*Solidarity is both a national and a human duty.*
The group agreed that universal or collective framing signals *Shared Values*, whereas direct encouragement addressed to a specific audience belongs to *Supportive Interactions*.

**Outcome of Quality Review.** All disagreement cases were documented in a shared reference sheet that now forms part of the dataset release materials. Each review cycle contributed to higher consistency and a deeper understanding of Arabic pragmatic nuance. The final manual incorporates explicit examples, counter-examples, and resolution notes derived from these analyses. This rigorous combination of human expertise, machine assistance, and documented quality control ensures the dataset's reliability, interpretive transparency, and reproducibility.

## 3.5. Comparing Manual and AI-Assisted Annotation

The annotation process itself yielded important methodological findings. Manual annotation excelled at identifying nuanced expressions like sarcasm and cultural references, offering deep contextual understanding. However, the process was time-intensive and mentally demanding, with a risk of annotator fatigue and subjective bias.

| Label | Manual Acc. | AI Acc. (Before Verif.) | AI Acc. (After Verif.) |
|---|---|---|---|
| Conflict | 93% | 85% | 90% |
| Supportive Interactions | 91% | 84% | 89% |
| Community Engagement | 94% | 87% | 91% |
| Shared Values | 90% | 82% | 88% |
| Resolution | 92% | 86% | 90% |

Table 2: Accuracy comparison for manual vs. AI-assisted annotation, evaluated against a 300-post gold-standard reference set derived from triple-coded posts with full annotator agreement.

AI-assisted annotation using ChatGPT offered remarkable speed and consistency. Its primary weakness was a tendency to struggle with ambiguity and to over-rely on surface-level keywords for category assignment, which could skew the dataset if left unverified. The integration of a human verification step proved critical: manual review corrected AI misclassifications and refined its output, significantly improving its accuracy.

As shown in Table 2, the final accuracy of the hybrid AI-assisted approach (88–91%) closely approached that of the purely manual process (90–94%), but was achieved at a fraction of the time and effort. This demonstrates the viability and effectiveness of a hybrid human-AI model for large-scale annotation of complex political discourse.

## 3.6. Data Availability

The *Cohesion-6K* dataset will be made available for research purposes via the following link.[1] The distributed version includes Arabic Facebook post texts, cohesion category labels for the annotated subset, aggregated engagement metrics, and the complete annotation guidelines used in the study.

In addition, the release package will provide access to a larger unannotated corpus of approximately 200,000 Arabic Facebook posts collected through CrowdTangle. This additional corpus is intended to support further analysis, model pretraining, and follow-up annotation projects.

Access to the resources is subject to a data-sharing agreement requiring (a) non-commercial academic use, (b) institutional ethics approval or equivalent waiver where applicable, and (c) a prohibition on attempts to re-identify individuals or original account holders. All materials are released under a Creative Commons Attribution Non-Commercial ShareAlike 4.0 (CC BY-NC-SA 4.0) License, consistent with the ethical considerations discussed in the ethics statement. Researchers are also advised to consult Meta's current Terms of Service regarding the reuse of derived social-media data.

[1] https://tinyurl.com/4ke5jwyw

## 4. Research Findings and Analysis

Our analysis of the 6,000 annotated Facebook posts reveals significant trends in how digital discourse on the Israeli-Palestinian conflict is shaped, amplified, and engaged with. The findings highlight the disproportionate visibility of conflict-driven content and provide insights into the mechanics of digital social cohesion.

### 4.1. Category Distribution and Dominance of Conflict

The initial analysis focused on the distribution of the five discourse labels across the dataset. The results show a clear dominance of content categorized as *Conflict*, which accounted for 40% of all posts. This was followed by *Community Engagement* (20%), while *Resolution* (15%), *Shared Values* (15%), and *Supportive Interactions* (10%) were considerably less frequent. This distribution underscores that divisive and antagonistic narratives are the most prevalent form of discourse in this online space, starkly overshadowing content aimed at reconciliation or mutual support (Bail, 2021).

### 4.2. Analysis of Engagement Patterns

| Discourse Type | Avg. Likes | Avg. Shares | Avg. Comments |
|---|---|---|---|
| Conflict | 5,630 | 1,980 | 4,120 |
| Resolution | 2,320 | 680 | 1,050 |
| Shared Values | 2,810 | 740 | 1,230 |
| Supportive Interactions | 2,220 | 600 | 910 |
| Community Engagement | 1,870 | 540 | 820 |

Table 3: Average engagement metrics across discourse types.

The most striking finding of this study is the dramatic disparity in user engagement across different discourse types. Posts categorized as *Conflict* received significantly higher levels of interaction (likes, shares, and comments) than any other category.

As shown in Table 3, conflict-oriented posts garnered on average 2.4 times more likes, 2.9 times more shares, and 3.9 times more comments than resolution-oriented posts. This pattern suggests that the attention economy of social media, driven by platform algorithms and user behavior, heavily favors emotionally charged and divisive content. To verify the statistical significance of this disparity, an independent-samples $t$-test was conducted comparing the engagement means for *Conflict* and *Resolution* posts. The result yielded $p < 0.01$, indicating a statistically significant difference in user interaction levels. This finding aligns with existing research indicating that content imbued with negative sentiment and divisive rhetoric is more effective at capturing user attention than neutral or conciliatory messages (Bail, 2021).

### 4.3. Thematic Trends in Digital Social Cohesion

| Page Type | Engagement (%) | Dominant Narrative |
|---|---|---|
| Pro-Palestinian Activist Pages | 42% | Critique of Israeli policies, calls for resistance. |
| Pro-Israeli Nationalist Pages | 35% | Defense of Israeli actions, nationalist rhetoric. |
| International News Organizations | 15% | Neutral or balanced reporting. |
| Peace and Dialogue Initiatives | 8% | Calls for reconciliation and co-existence. |

Table 4: Ideological distribution of dominant Facebook groups and pages.

It should be noted that the ideological categories in Table 4 reflect a qualitative characterization of dominant pages by their observed narrative orientation, and are distinct from the source-type stratification used for dataset sampling in Table 1.

A qualitative analysis revealed distinct thematic patterns within each category. Conflict narratives frequently centered on historical grievances, territorial claims, and the reinforcement of in-group and out-group identities. Resolution-oriented content, while less common, typically advocated for interfaith dialogue, ceasefires, and humanitarian aid. Community Engagement posts focused on mobilization efforts such as protests, advocacy, and fundraising, peaking during critical events like the Sheikh Jarrah evictions.

This analysis also highlighted the formation of insulated discourse ecosystems, commonly referred to as echo chambers. Users immersed in conflict-driven content were found to engage predominantly with pages and groups that mirrored their existing views, thereby diminishing opportunities for inter-group dialogue. The ideological distribution of dominant pages further illustrates this trend.

As shown in Table 4, pages with strong ideological stances (pro-Palestinian and pro-Israeli) captured a combined 77% of total engagement, while pages dedicated to peace and dialogue initiatives received only 8%. This demonstrates the structural challenges faced by conversations aimed at promoting peace and problem-solving.

# 5. Discussion

The findings from *Cohesion-6K* carry several implications for how we understand Arabic digital discourse, platform design, and computational approaches to social cohesion. The consistent amplification of conflict-driven content across likes, shares, and comments points to a reinforcing dynamic between user behavior and algorithmic curation: emotionally charged content generates more interaction, which in turn increases its visibility. This cycle makes it structurally difficult for resolution-oriented or community-focused content to reach comparable audiences.

From a platform governance perspective, these findings raise questions about how recommendation systems might be adjusted to surface cohesion-oriented content without suppressing organic user expression. The dataset can serve as a benchmark for evaluating such interventions computationally.

From an Arabic NLP perspective, *Cohesion-6K* fills a gap in discourse-level resources for a language that remains underrepresented in the field. The five-category taxonomy captures pragmatic and functional dimensions of language use that go beyond sentiment polarity, offering a richer signal for downstream models. The annotation challenges documented in Section 3.4.1, particularly around sarcasm, religious expressions, and code-switched text, also provide a practical foundation for future work on Arabic pragmatic understanding.

It should be noted that the single-label design of the annotation scheme, while enabling cleaner analysis, limits the ability to capture posts that simultaneously express multiple discourse functions. Future iterations of the dataset could explore multi-label annotation to address this limitation.

# 6. Conclusion and Future Work

This study introduced *Cohesion-6K*, a large-scale Arabic dataset designed to examine how online discourse reflects and influences social cohesion in politically polarized contexts. Through a combination of manual and AI-assisted annotation, the dataset captures fine-grained distinctions between conflict-oriented and cohesion-oriented narratives across 6,000 Facebook posts related to the Israeli Occupation of Palestine.

Our analysis revealed that conflict-driven content dominates online discussion, receiving disproportionately higher engagement than posts that advocate dialogue, community engagement, or shared values. This imbalance suggests that the current social media attention economy structurally amplifies divisive discourse, which may hinder collective understanding and digital resilience. The *Cohesion-6K* dataset provides a reproducible foundation for analyzing these mechanisms and developing computational tools to detect, visualize, and eventually counteract polarization dynamics in Arabic social media.

Future research can expand this work along several dimensions. Cross-platform analyses would help test the generalizability of these findings across different engagement architectures such as X, Instagram, and TikTok. Longitudinal studies could capture how narrative trends evolve over time in response to political and humanitarian events. From a computational perspective, the dataset can serve as training material for models that detect discourse types beyond sentiment, such as moral framing, solidarity, or civic engagement, supporting the growth of culturally grounded Arabic NLP resources.

# 7. Limitations

While this study contributes an original resource and analysis, several limitations should be acknowledged.

### Platform and Scope

The dataset focuses exclusively on Facebook, whose demographic and algorithmic structures differ from other social platforms. Results may therefore not generalize directly to environments such as TikTok or X, where visual content and brevity shape discourse differently. Future comparative research across multiple platforms could provide a more comprehensive view of digital cohesion in Arabic social media.

### Temporal and Contextual Constraints

The dataset represents a specific historical window and captures reactions to recent escalations but does not track long-term discourse shifts. Political and social events evolve quickly, and public sentiment can change in response to new developments. Extending this dataset into a longitudinal corpus would enable dynamic modeling of how narratives of conflict and resolution evolve over time.

### Annotation Scope and Single-Label Design

The single-label annotation scheme, while enabling straightforward thematic analysis, limits visibility into posts that carry multiple simultaneous discourse functions. The use of a single label per post is an acknowledged simplification of the inherent complexity of social discourse, and other researchers working with similar data may arrive at different or expanded categorizations. The annotation guidelines will be released alongside the dataset to support transparency and future comparative work.

### Annotation Bias and Interpretation

Despite extensive training, clear guidelines, and a high inter-annotator agreement (Cohen's $\kappa = 0.85$), annotation inherently involves subjective judgment. Cultural, political, or linguistic perspectives may influence how annotators interpret complex posts, particularly those involving irony or coded language. Continuous revision of the annotation manual and inclusion of diverse annotator backgrounds can further enhance neutrality and inclusiveness in future iterations.

## 8. Ethical Considerations

Working with politically sensitive, user-generated data requires careful adherence to ethical standards in data collection, processing, and dissemination. This study followed established best practices for digital research ethics and transparency.

### Data Privacy and Anonymization

All data were obtained through the CrowdTangle platform, which provides access only to publicly available posts from verified or public Facebook pages. No private content, direct user data, or personal identifiers were collected. All metadata were anonymized before analysis and stored securely. The dataset will be released for academic research under a Creative Commons Attribution NonCommercial ShareAlike 4.0 license.

### Minimizing Annotator Bias and Emotional Harm

Given the emotionally charged nature of the content, annotators were trained to approach the material with cultural sensitivity and methodological neutrality. Calibration sessions emphasized reflexivity, meaning awareness of how one's own perspective may influence interpretation. Annotators also had the option to skip or discuss posts perceived as distressing, ensuring both data quality and participant well-being.

### Misinformation and Responsible Reporting

The dataset includes posts that may contain misinformation or partisan framing. Such posts were not removed but labeled and analyzed objectively to maintain representational integrity. In reporting findings, care was taken to avoid reinforcing divisive narratives or amplifying harmful content. All interpretations prioritize empirical accuracy over ideological alignment.

### Transparency and Reproducibility

All annotation guidelines, documentation, and aggregated data will be shared publicly upon publication to enable verification and reuse by the research community. This commitment to openness supports cumulative progress in the study of Arabic digital discourse and computational social science.

## Acknowledgments

This work was made possible by the National Priorities Research Program grant NPRP14C-0916-210015 from the Qatar National Research Fund (QNRF), part of the Qatar Research, Development and Innovation Council (QRDI).